\def\year{2017}\relax
\documentclass[letterpaper]{article}
\usepackage{aaai17}
\usepackage{times}
\usepackage{helvet}
\usepackage{courier}
\usepackage{latexsym}
\usepackage{amsmath}
\usepackage{amssymb}
\usepackage{algorithm}
\usepackage{algpseudocode}
\usepackage{pifont}
\usepackage{caption}
\usepackage{subcaption}
\usepackage{graphicx}
\usepackage{booktabs}
\usepackage{float}
\usepackage{natbib}
\usepackage{pgfplots,xcolor}

\author{
Volkan Cirik, Eduard Hovy, Louis-Philippe Morency\\
Department of Computer Science\\
Language Technologies Institute \\
Carnegie Mellon University\\
Pittsburgh, PA 15213 \\
\texttt{\{vcirik,hovy,morency\}@cs.cmu.edu} \\
}

\begin{document}
\title{Visualizing and Understanding Curriculum Learning \\for Long Short-Term Memory Networks}

\maketitle
\begin{abstract}

Curriculum Learning emphasizes the order of training instances in a computational learning setup.
The core hypothesis is that simpler instances should be learned early as building blocks to learn more complex ones.
Despite its usefulness, it is still unknown how exactly the internal representation of models are affected by curriculum learning.
In this paper, we study the effect of curriculum learning on Long Short-Term Memory  (LSTM) networks, which have shown strong competency in many Natural Language Processing (NLP) problems.

Our experiments on sentiment analysis task and a synthetic task similar to sequence prediction tasks in NLP show that curriculum learning has a positive effect on the LSTM's internal states by biasing the model towards building constructive representations i.e. the internal representation at the previous timesteps are used as building blocks for the final prediction.
We also find that smaller models significantly improves when they are trained with curriculum learning.
Lastly, we show that curriculum learning helps more when the amount of training data is limited.

\end{abstract}

\section{Introduction} \label{sec:introduction}

Inspired by the human learning process, Curriculum Learning \citep{elman1993learning,bengio2009curriculum} is an algorithm that emphasizes the order of training instances in a computational learning setup.
The main idea is that learning easy instances first could be helpful for learning more complex ones later in the training.
The first algorithm proposed by \citet{bengio2009curriculum}, which we refer as \textit{one-pass} curriculum, creates disjoint sets of training examples ordered by the complexity and used separately during training.
The second algorithm called \textit{baby step} curriculum uses an incremental approach where groups of more complex examples are incrementally added to the training set \citep{spitkovsky2010baby}.
These curriculum learning regimens were shown to improve performance in some Natural Language Processing and Computer Vision tasks \citep{pentina2015curriculum,spitkovsky2010baby}.

Despite its usefulness, it is still unknown how exactly computational models are affected internally by curriculum learning.
An example of computational model particularly relevant to Natural Language Processing is Long Short-Term Memory (LSTM) network \citep{hochreiter1997long}.
LSTM networks have shown competitive performance in several domains such as handwriting recognition \citep{graves2009novel} and parsing \citep{vinyals2015grammar}.
Surprisingly, curriculum learning has not been studied in the context of LSTM networks to our knowledge.
Detailed visualizations and analyses of curriculum learning regimens with LSTM will allow us to better understand how models are affected and provides us insights when to use these regimens. Knowing how curriculum learning works, we can design new extensions and understand the nature of tasks most suited for these learning regimens.

In this paper, we study the effect of curriculum learning on LSTM networks. 
We created experiments to directly compare two curriculum learning regimens, \textit{one-pass} and \textit{baby step}, with two baseline approaches that include the conventional technique of randomly ordering the training samples.
We use two benchmarks for our analyses.
First, a synthetic task is designed which is similar to several Natural Language Processing tasks where a sequence of symbols are observed and a particular function (e.g. analogous to a linguistic or a semantic phenomenon) is aimed to be learned.
Second, we use sentiment analysis where the polarity of subjective opinions is classified -- a fundamental task in Natural Language Processing.
As mentioned previously, this is the first work studying LSTM networks on sentiment analysis with curriculum learning to our knowledge.

Our visualizations and analyses on these two sequence tasks are designed to study three main factors. 
First, we compare the four learning regimens on how the LSTM network's internal representations change as the final prediction is computed.
To this end, we simply decode the representations at intermediate steps.
This analysis helps us understand how a model handles the task with the help of curriculum learning.
Second, we investigate how the performance of models with different complexities are affected by curriculum learning.
Smaller yet accurate models are crucial in limited resource settings.
Third, we study how the performance of curriculum learning changes in low-resource setups.
This analysis provides us a valuable information considering low-resource scenarios are common in several data-driven learning domains such as Natural Language Processing.
\section{Related Work} \label{sec:related}
We review a list of topics related to our work in the context of curriculum learning (CL), analysis of neural networks and sentiment analysis with neural networks.

\paragraph{Curriculum Learning.} Motivated by children's language learning, \citet{elman1993learning} studies the effect of learning regimen on a synthetic grammar task.
He shows that a Recurrent Neural Network (RNN) is able to learn a grammar when training data is presented from simple to complex order and fails to do so when the order is random. \citet{bengio2009curriculum} investigate CL from an optimization perspective.
Their experiments on synthetic vision and word representation learning tasks show that CL results in better generalization and faster learning.
\citet{spitkovsky2010baby} apply a CL strategy to learn an unsupervised parser for sentences of length $k$ and initialize the next parser for sentences of length $k+1$ with the previously learned one.
They show that learning a hybrid model using the parsers learned for each sentence lengths achieves a significant improvement over a baseline model.
\citet{pentina2015curriculum} investigate the CL in a multi-task learning setup and propose a model to learn the order of multiple tasks.
Their experiments on a set of vision tasks show that learning tasks  sequentially is better than learning them jointly.
\citet{jiang2015self} provide a general framework for CL and Self-Paced Learning where model picks which instances to train based on a simplicity metric.
The proposed framework is able to combine prior knowledge of curriculum with Self-Paced Learning in the learning objective.

\paragraph{Long Short-Term Memory Networks.} LSTM networks are a variant of RNNs \citep{elman1990finding} capable of storing information and propagating loss over long distance.
Using a gating mechanism by controlling the information flow into the internal representation, it is possible to avoid the problems of training RNNs \citep{bengio1994learning,pascanu2012difficulty}.
Several architectural variants have been proposed to improve the basic model \citep{cho2014properties,chung2015gated,yao2015depth,kalchbrenner2015grid,dyer2015transition,grefenstette2015learning}. 

\paragraph{Visualization of Neural Networks.} Although many of the neural network studies provide quantitative analysis, there are few qualitative analyses of neural networks.
\citet{zeiler2014visualizing} visualize the feature maps of a Convolutional Neural Network (CNN) \citep{lecun1998gradient}.
They show that feature maps at different layers show sensitivity to different shapes, textures, and objects.
Similarly, \citet{karpathy2015visualizing} analyze LSTM on character level language modeling.
Their analysis shows that deeper models with gating mechanisms achieve better results.
They show that some cells in LSTM  learn to detect patterns and how RNNs learn to generalize to longer sequences.
More recently, \citet{li2016} use visualization to show how neural network models handle several linguistics phenomena such as compositionality and negation using sentiment analysis and sequence auto-encoding.

\textbf{Synthetic Tasks.} Since the early days of neural networks, synthetic tasks were used to test the capabilities of the models \citep{fahlman1991recurrent} and often serve as unit tests for machine learning models \citep{weston2015towards}.
Similar to the first work on LSTM \citep{hochreiter1997long}, many of the contemporary neural network models \citep{graves2014neural,kurach2015neural,sukhbaatar2015end,vinyals2015pointer} use synthetic tasks to compare and contrast several architectures.
Inspired by these studies, we also use a synthetic task as one of our tasks to understand the effect of CL on LSTMs.

\paragraph{Sentiment Analysis with Neural Networks.}
Several approaches have been proposed to solve sentiment analysis using neural networks.
\citet{socher2013recursive} propose Recursive Neural Networks to exploit the syntactic structure of a sentence.
A number of extensions of this model have been proposed in the context of sentiment analysis \citep{irsoy2014deep,tai2015improved}.
Other proposed approaches use CNN \citep{kalchbrenner2014convolutional,kim2014convolutional} and the averaging of word vector models \citep{iyyer2015deep,le2014distributed}.

To our knowledge, this work is the first to study how the internal representation of LSTM change in a curriculum learning setup.

\section{Curriculum Learning Regimens} \label{sec:regimens}

Curriculum learning emphasizes the order of training instances, prioritizing simpler instances before the more complex ones. In this section, we describe two curriculum learning regimens: one-pass curriculum originally proposed by \citet{bengio2009curriculum} and baby step curriculum from \citet{spitkovsky2010baby}. For both regimens, we develop the curriculum $\mathcal{C}$ using the same strategy proposed by \citet{spitkovsky2010baby} who assume that shorter sequences are easier to learn. 

The following subsections are describing the two curriculum learning regimens as well as two baseline learning regimens. 

\subsection{One-Pass Curriculum}\label{ssec:onepass}

\citet{bengio2009curriculum} propose to use a dataset with simpler instances in the first phase of the training.
After some number of iterations, they switch to harder target dataset. 
The intuition is that after some training on simpler data, the model $\mathnormal{M}$ is ready to handle the harder target data. 
Here, we name this regimen One-Pass curriculum (see Algorithm~\ref{alg:onepass}).
The training data $\mathcal{D}$ is sorted by a curriculum $\mathcal{C}$ and distributed into $k$ number of buckets. 
The training starts with the easiest bucket.
Unlike the previous work \citep{bengio2009curriculum}, we use early stopping -- training stops for the bucket when the loss or task's accuracy criteria on held-out set do not get any better for $p$ number of epochs.
Afterward, the next bucket is being used and trained in the same way.
The whole training is stopped after all buckets are used.
Note that the model uses each bucket only one time for the training, hence the name.

\begin{algorithm}
\caption{One-Pass Curriculum}\label{alg:onepass}
\begin{algorithmic}[1]
\Procedure{OP-Curriculum}{$\mathnormal{M}$,$\mathcal{D}$, $\mathcal{C}$}
\State $\mathcal{D}'$ = sort($\mathcal{D}$, $\mathcal{C}$)
\State $\{\mathcal{D}^1,\mathcal{D}^2,...,\mathcal{D}^k\} = \mathcal{D}' $ where $\mathcal{C}(d_a) < \mathcal{C}(d_b)$ $d_a \in D^i$ , $d_b \in D^j$, $\forall i<j$

\For{ $s$ = 1...$k$ }
\While{not converged for $p$ epochs}
	\State train($\mathnormal{M}$, $\mathcal{D}^{s}$)    
\EndWhile
\EndFor
\EndProcedure
\end{algorithmic}
\end{algorithm}

\subsection{Baby Steps Curriculum}\label{ssec:baby}

The intuition behind Baby Steps curriculum \citep{bengio2009curriculum,spitkovsky2010baby} is that simpler instances in the training data should not be discarded, instead, the complexity of the training data should be increased.
After distributing data into buckets based on a curriculum, training starts with the easiest bucket.
When the loss or task's accuracy criteria on a held-out set do not get any better for $p$ number of epochs,  the next bucket and the current data bucket are merged.
The whole training is stopped after all buckets are used (see Algorithm~\ref{alg:baby}). 

\begin{algorithm}
\caption{Baby Steps Curriculum}\label{alg:baby}
\begin{algorithmic}[1]
\Procedure{BS-Curriculum}{$\mathnormal{M}$,$\mathcal{D}$, $\mathcal{C}$}
\State $\mathcal{D}'$ = sort($\mathcal{D}$, $\mathcal{C}$)
\State $\{\mathcal{D}^1,\mathcal{D}^2,...,\mathcal{D}^k\} = \mathcal{D}' $ where $\mathcal{C}(d_a) < \mathcal{C}(d_b)$ $d_a \in D^i$ , $d_b \in D^j$, $\forall i<j$
\State $\mathcal{D}^{train} = \O $
\For{ $s$ = 1...$k$ }
\State $\mathcal{D}^{train} = \mathcal{D}^{train} \cup \mathcal{D}^s $
\While{not converged for $p$ epochs}
	\State train($\mathnormal{M}$, $\mathcal{D}^{train}$)    
\EndWhile
\EndFor
\EndProcedure
\end{algorithmic}
\end{algorithm}

\begin{table*}[t]
\centering
\caption{Probing of the LSTM model at intermediate timesteps for the Digit Sum dataset. Left column is the input and underlined digit emphasizes the last input digit. Ground Truth is the running sum up to that point. Predicted values by the LSTM models are in prediction column. The number in parantheses are the standard deviation. The intermediate representation of Baby Step curriculum model is closer to running sum of the input sequence. }
\label{tab:ss-analyze}
\scalebox{0.80}{
\begin{tabular}{ll|l|l|l|l}
\hline
                                   &             & Baby Steps Curriculum & One-Pass Curriculum & Sorted & No-CL        \\ \hline
Input Sequence                     & Ground Truth & Prediction            & Prediction          & Prediction        & Prediction   \\ \hline
\underline{1}                    & 1           & 0.23                  & 0.00                & 0.86              & 0.88 (0.20)  \\
1\underline{0}                   & 1           & 1.03                  & 0.00                & 1.32              & 1.10 (0.28)  \\
10\underline{9}                  & 10          & 10.27                 & 0.00                & 10.59             & 10.20 (0.78) \\
109\underline{1}                 & 11          & 11.40                 & 0.00                & 11.29             & 10.88 (0.95) \\
1091\underline{7}                & 18          & 18.62                 & 0.00                & 19.30             & 17.89 (1.49) \\
10917\underline{3}               & 21          & 21.54                 & 0.00                & 22.99             & 20.84 (1.94) \\
109173\underline{5}              & 26          & 26.77                 & 4.29                & 29.34             & 25.90 (2.39) \\
1091735\underline{6}             & 32          & 32.82                 & 14.19               & 37.21             & 32.00 (2.87) \\
10917356\underline{7}            & 39          & 40.56                 & 29.28               & 46.38             & 39.12 (3.12) \\
109173567\underline{0}           & 39          & 40.53                 & 32.20               & 48.01             & 39.26 (3.00) \\
1091735670\underline{6}          & 45          & 46.73                 & 52.86               & 55.50             & 45.95 (3.14) \\
10917356706\underline{4}         & 49          & 50.96                 & 67.70               & 60.58             & 50.32 (3.09) \\
109173567064\underline{2}        & 51          & 52.82                 & 74.59               & 63.03             & 52.6 (2.94)  \\
1091735670642\underline{8}       & 59          & 61.01                 & 83.31               & 70.91             & 61.11 (2.87) \\
10917356706428\underline{6}      & 65          & 67.78                 & 87.42               & 76.54             & 67.97 (2.77) \\
109173567064286\underline{1}     & 66          & 69.27                 & 83.43               & 76.46             & 69.33 (2.57) \\
1091735670642861\underline{4}    & 70          & 72.32                 & 82.60               & 78.88             & 73.05 (2.36) \\
10917356706428614\underline{5}   & 75          & 76.77                 & 83.34               & 81.56             & 77.67 (2.27) \\
109173567064286145\underline{1}  & 76          & 78.57                 & 80.42               & 80.44             & 78.07 (2.19) \\
1091735670642861451\underline{6} & 82          & 83.05                 & 82.52               & 83.68             & 83.36 (2.03) \\ \hline
\end{tabular}
}
\vspace{-10pt}
\end{table*}

\subsection{Baseline Regimens}\label{ssec:baselines}
The first baseline, named \textbf{No-CL}, is the common practice of shuffling the training data. For a neural network like our LSTM models, this means that training is performed as usual where one epoch sees all the training set in random order. For all experiments described in the following section, models learned with the No-CL regiment are trained 10 times\footnote{Note that this favors No-CL due to lower variance in results.}, to get a proper average performance.

The second baseline, named \textbf{Sorted}, also sees all the data at each epoch but the ordering of the training instances is based on the curriculum $\mathcal{C}$. This is a simplification of the two CL regimens presented in the previous subsections since we are not partitioning the data based on its complexity (i.e., based on the curriculum). We are simply reordering the training set.
A comparison between the No-CL and Sorted baselines will allow us to study the importance of training instance ordering. 

\section{Experiments} \label{sec:experiments}

The main goal of our experiments is to better understand how a computational model, specifically LSTM networks, are affected internally by CL.
We aim to observe (1) the effect of CL on the internal model representations, (2) how the number of model parameters affect the performance of CL, and (3) how the amount of data size change the contribution of CL.

The following subsections present LSTM network and our experimental probing methodology to analyze the LSTM’s internal representations at different stages in the sequence modeling process. 

\begin{figure*}[t]
\vspace{-10pt}
  \centering
      \begin{subfigure}{0.49\textwidth}
    \centering
      \includegraphics[width=0.99\textwidth]{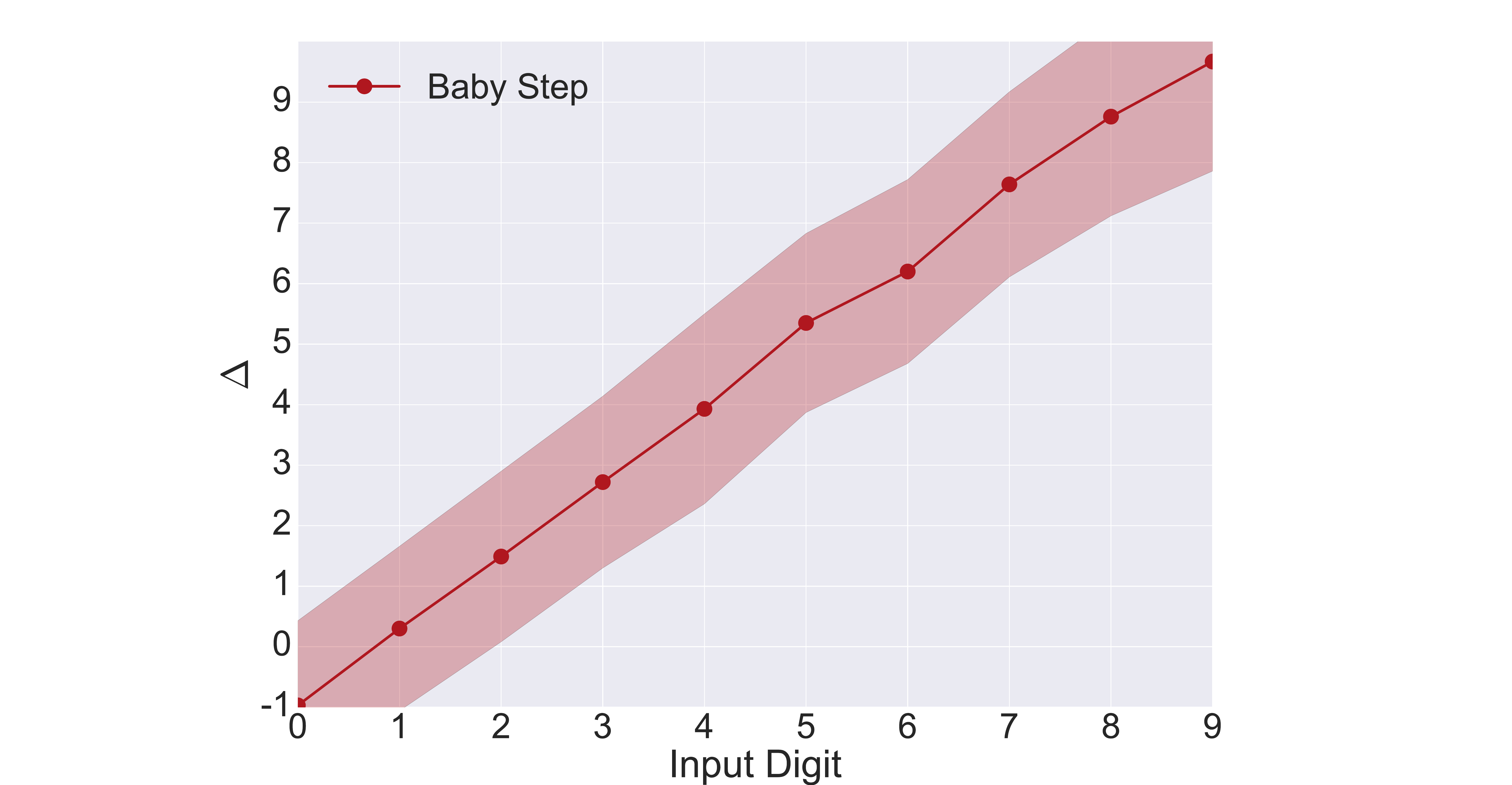}
  \end{subfigure}
        \begin{subfigure}{0.49\textwidth}
    \centering
      \includegraphics[width=0.99\textwidth]{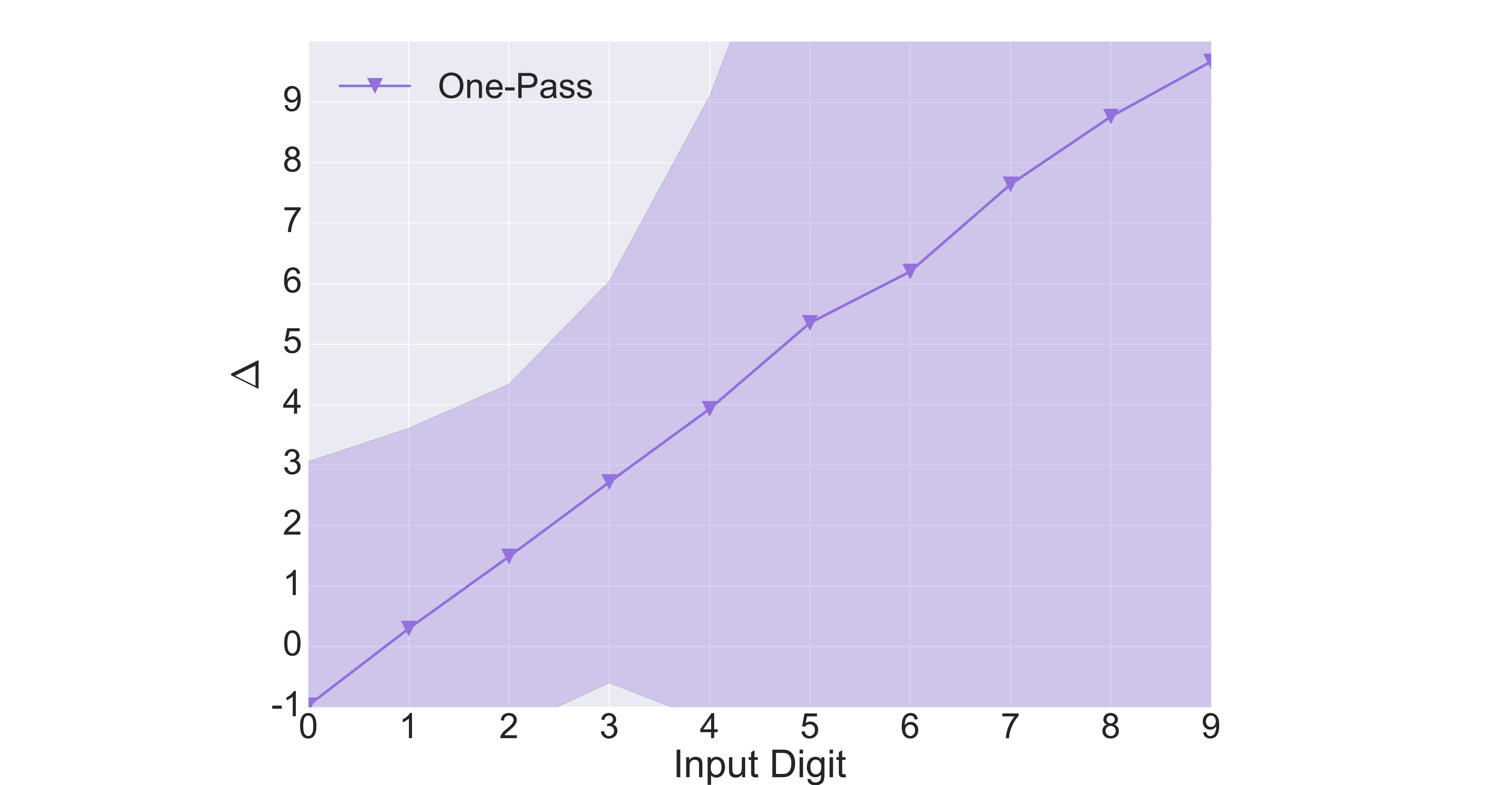}
  \end{subfigure}
      \begin{subfigure}{0.49\textwidth}
    \centering
      \includegraphics[width=0.99\textwidth]{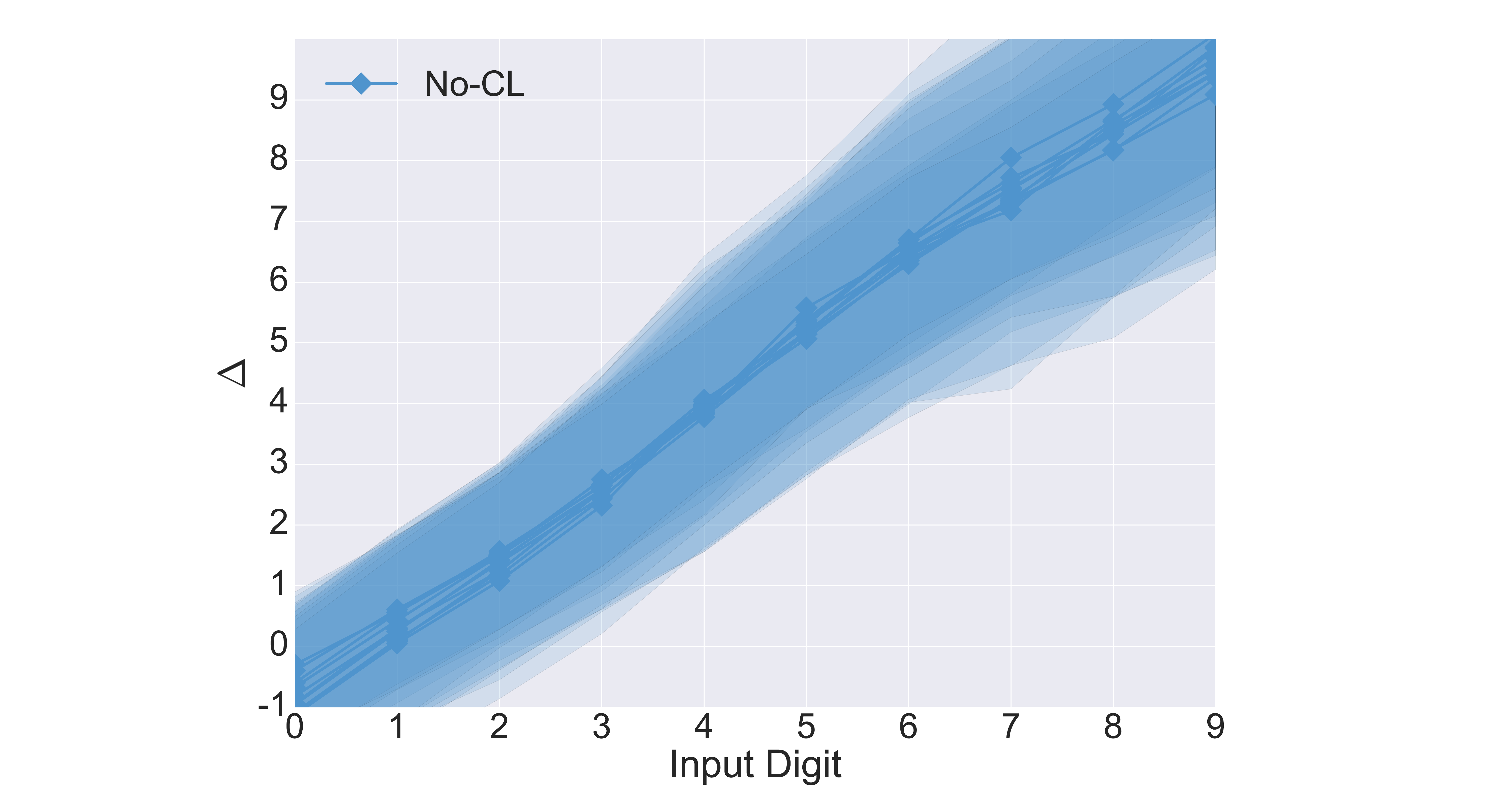}
  \end{subfigure}
        \begin{subfigure}{0.49\textwidth}
    \centering
      \includegraphics[width=0.99\textwidth]{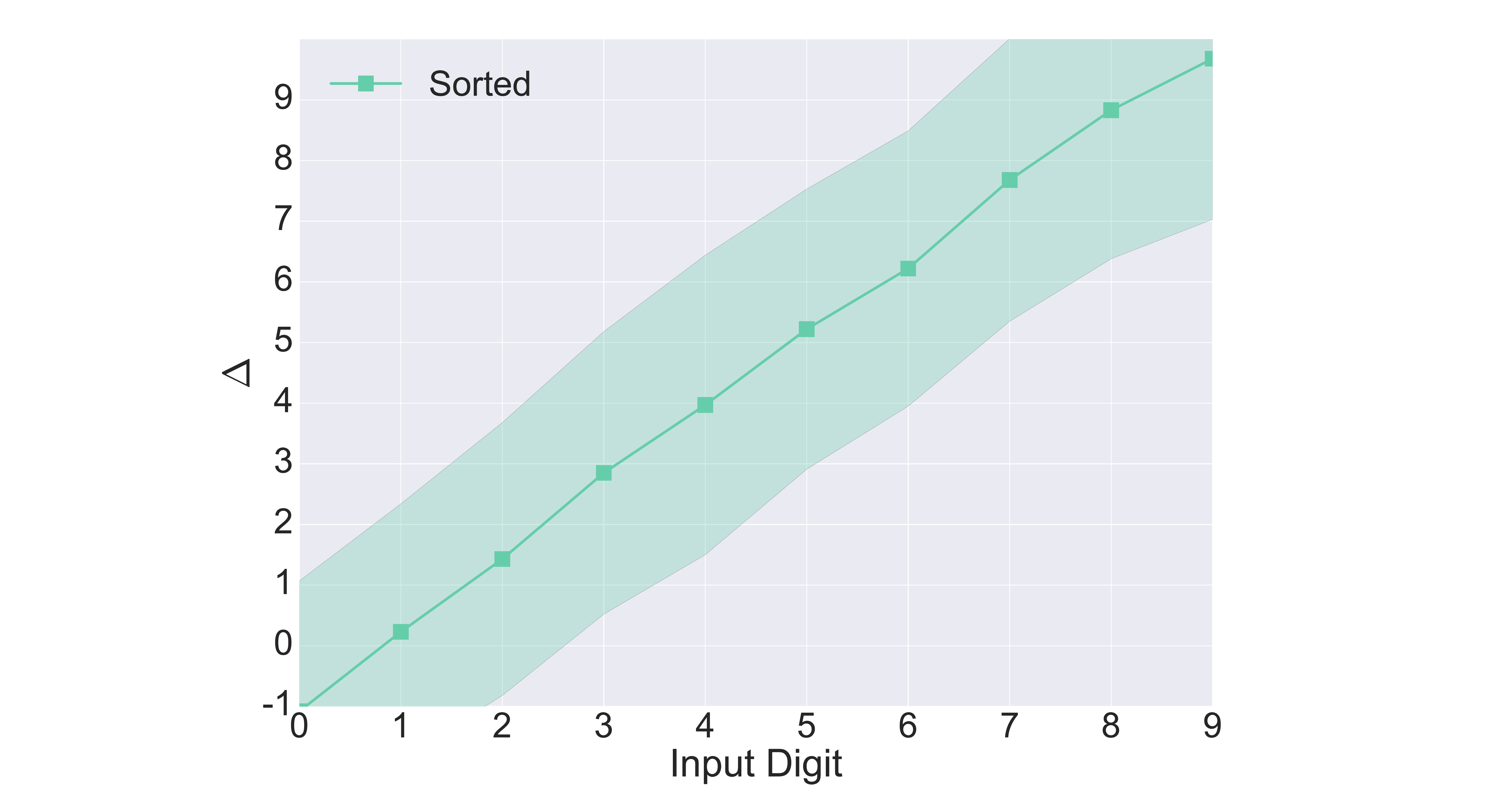}
  \end{subfigure} 
  \caption{Visualization of Input Digit (input digit at time $t$) vs $\Delta$ (the average difference between predictions at $t$ and $t-1$). Shaded areas represent the variance.
We expect $\Delta$ to follow the input digit.
Model trained with Baby Step curriculum shows the least variance the input digit.
Note that for No-CL, we plot all 10 runs (with different seeds).
}\label{fig:delta}
\vspace{-10pt}
\end{figure*}

\subsection{LSTM}\label{ssec:prob}

We now describe LSTM networks.
Let $x_1,..,x_T$ be a sequence of one-hot coded sequence of symbols of length $T$. At each time step the LSTM updates its cells as follows:

\begin{align}
   \begin{pmatrix}i \\ f \\ o \\ m \end{pmatrix} & = \begin{pmatrix} \text{sigm} \\ \text{sigm} \\ \text{sigm} \\ \text{tanh} \end{pmatrix} W^t \begin{pmatrix} x_t W^{e} \\ h_{t-1} \end{pmatrix} \\
   c_t & =  f \odot c_{t-1} + i \odot m \\
h_t & =  o \odot \text{tanh}(c_t)   
\vspace{-10pt}
\end{align}

In above equations, $W^t$ is a [$4n \times 2n$] matrix to calculate gate weights and new memory information $g$. $W^e$ is an embedding matrix for symbols.
At time $t$, the sigmoid ($\text{sigm}$) and tanh ($\text{tanh}$) non-linearities are applied element-wise to the embedding representation of input $x_t W^{e}$ and the output of the network from the previous time step $h_{t-1}$.
Vectors $i,f,o \in \mathbb{R}^n$ are binary gates controlling the input, forget and output gates respectively.
The vector $m\in \mathbb{R}^n$ additively modifies the cell $c_t$.

We use the final hidden representation $h_T$ of the LSTM to prediction with a projection matrix $W^p$.
In the case of regression, we use $\text{relu}(W^p h^T)$  where $W^p$ is [$1 \times n$] and $\text{relu}$ is rectified non-linearity.
For classification, we predict one of $k$ class labels using $\text{softmax}(\text{relu}(W^p h^T)))$   where $W^p$ is [$k \times n$] and $\text{softmax}$ is the softmax function.

\subsection{Probing Internal States of LSTM}\label{ssec:prob}

We aim to observe how the use of the internal representation of LSTM at intermediate steps changes depending on the learning regimen.

Each internal representation $h_t$ is probed using the $\text{relu}(W^p h^t)$ functions learned for regression or the $\text{softmax}(\text{relu}(W^p h^t))$ function learned for classification.
By moving these probes along the sequences, we can study the intermediate representation at each time $t$.

\subsection{Digit Sum}\label{ssec:ss}

We aim to simulate a low-resource sequence regression problem considering many of the NLP tasks only have a few thousand annotated samples.
To this end, the Digit Sum task is posed as follows.
Given a sequence of symbols of digits, the model is expected to predict the sum of digits.
For instance given a sequence "5 0 2 4 6" the expected output is 17.

Digit Sum task has similarities with our second sequence task, sentiment analysis, where digits are analogous to the word tokens in the natural language text and the summation is analogous to the subjective position of a sequence on a topic.
Our two evaluation tasks also have some interesting differences which allow evaluating a broader range of sequence learning tasks. In sentiment analysis, the order of words makes a difference whereas, in the Digit Sum, the order of digits does not change the expected answer. Secondly, the learning setup is a classification of polarity levels for sentiment analysis whereas it is a regression for the Digit Sum.

\noindent \textbf{Dataset Details.} We define the evaluation task in the Digit Sum dataset as the summation of 20 digits, a typical length of sentences in natural language.
Both the validation and testing sets contain 200 sequences of 20 digits randomly generated.
The training set consist of 1000 sequences each from length 2 to 20, allowing to develop the curriculum automatically following \citet{spitkovsky2010baby} procedure. This results in a dataset of size 19K instances\footnote{We experimented with 10x smaller dataset size and observed very similar results. We do not report these due to limited space.}.

\noindent \textbf{Experimental Details.} We used LSTM with hidden units of $2,4,8,...,512$ without peephole connections. 
For all configurations the size of digit embeddings and hidden units are the same. 
We use RMSprop \citep{dauphin2015rmsprop} with learning rate 0.001 and decay rate of 0.9 with minibatches of size 128. The patience parameter $p$ for early stopping is 10.
We use Dropout \citep{srivastava2014dropout} of rate in range \{0,0.25,0.5\} as suggested for LSTMs by \citet{gal2015theoretically}.

\subsection{Results}

\noindent \textbf{Probing Internal Model Representations.} We analyze the behavior of the model by using the intermediate representations during processing of a sequence.
As we discussed previously, we feed the hidden representations of each digit to the regression node to predict at each timestep of the input.
Table~\ref{tab:ss-analyze} shows the input sequence, ground truth, and predictions of the best models for each learning regimen based on validation loss.
The prediction of the model trained with One-Pass curriculum and Sorted Baseline shows no correlation with the running sum of the digits. 
The Baby Step curriculum model is able to predict similar values to running sum.

\begin{figure*}[t]
\centering
\begin{center}
\centerline{\includegraphics[width=0.805\textwidth,height=0.805\textheight,keepaspectratio]{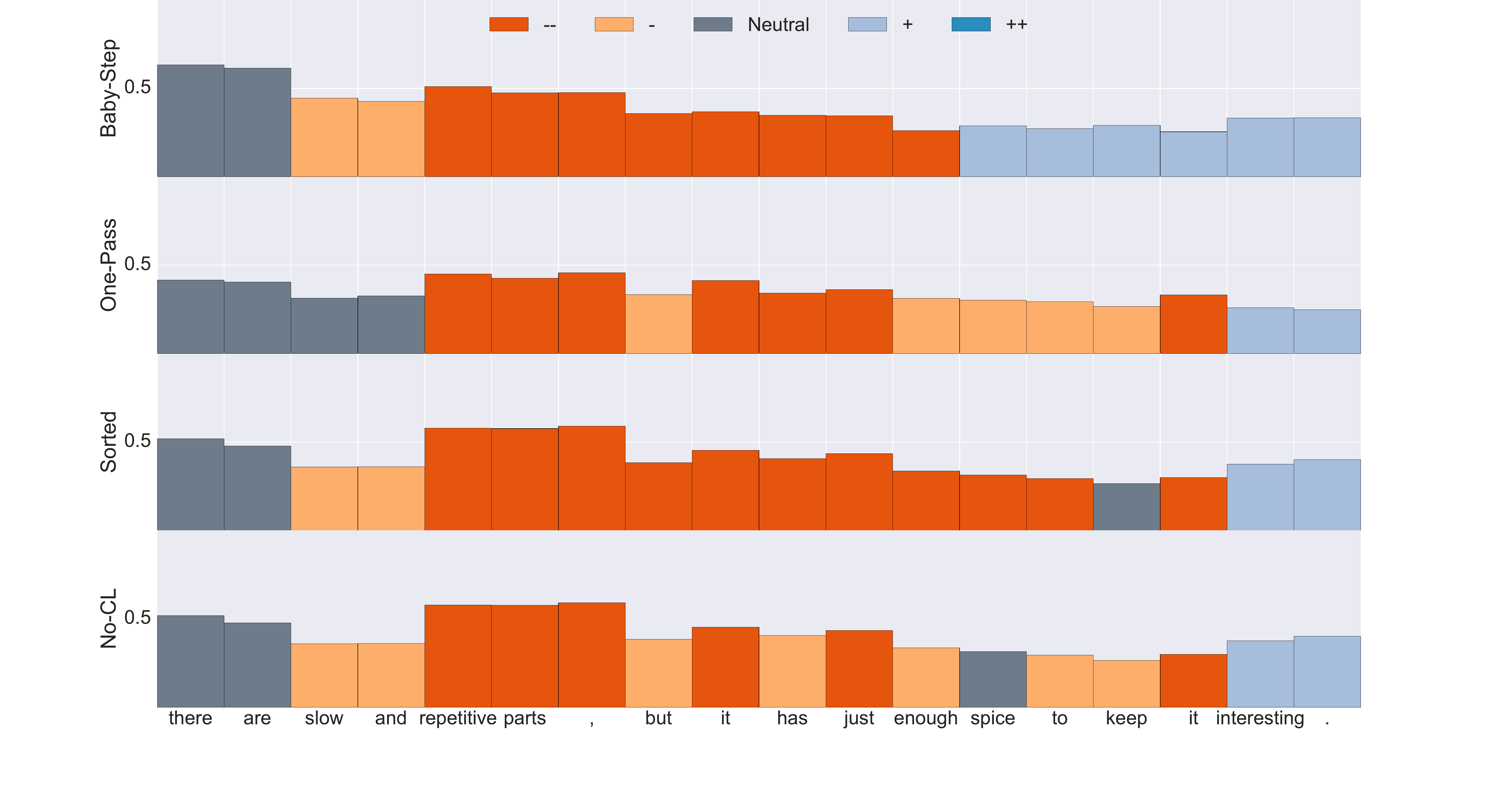}}
\caption{Predictions at intermediate tokens. Colors represent the polarity and heights represent the prediction probability. Baby Step curriculum shows a consistent behavior : only after observing a positive sub-phrase flips the prediction.}
\label{fig:sa-conj}
\end{center}
\vspace{-30pt}
\end{figure*}
A sequence model can learn to solve this task in numerous ways such as memorizing sequences due to overfitting, using a count table of the digits, or doing a running sum at each time step.
To analyze this, we report the average differences between successive predictions (we call it $\Delta$) and the last input digit(see Figure~\ref{fig:delta}).
At each timestep, the model trained with Baby Step curriculum updates the hidden representation such that it correlates with the sum of digits observed up to that point.
It is also interesting to observe that Baby Step curriculum shows better variance than the average of 10 random starts (No-CL).
We emphasize that models are provided with the same sequences for training, yet, the learning regiments results in different models.

\begin{figure}[h]
\centering
\centerline{\includegraphics[width=\columnwidth]{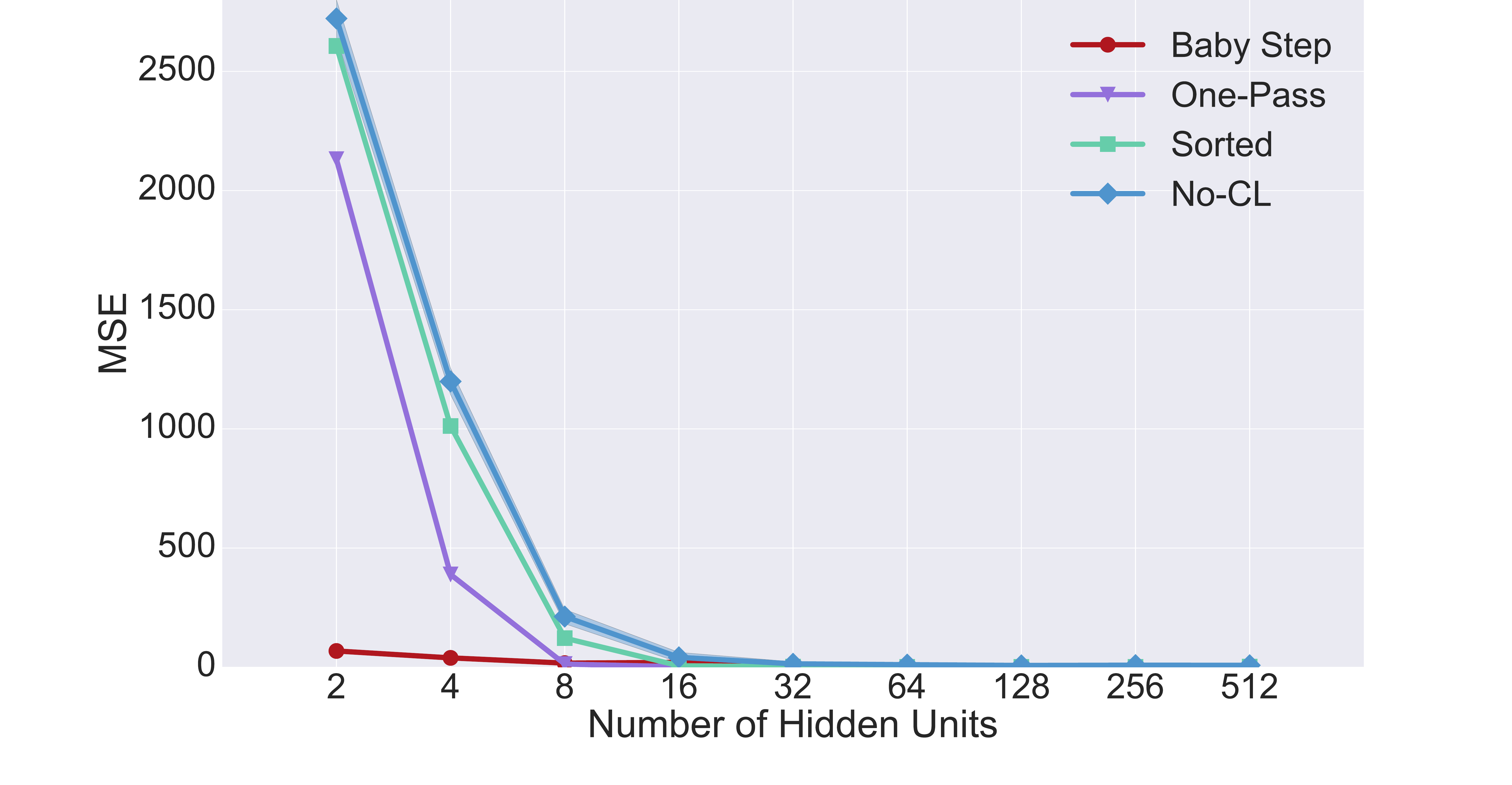}}
\vspace{-3pt}
\caption{Mean Squared Error vs the number of units of LSTM. With much smaller model, Baby Step curriculum achieves the best results. The other model requires the right complexity to achieve comparable results.}
\label{fig:unit-mse}
\vspace{-10pt}
\end{figure}

\noindent \textbf{Effects on Models With Different Complexities.} Our next experiment studies the effect of learning regimens on models with different complexities. Figure~\ref{fig:unit-mse} shows the Mean Squared Loss (MSE) results for the Digit Sum with LSTMs with varying hidden unit sizes.
Baby Step curriculum achieves consistently better results even if the model has much fewer parameters.
Other regimens require the model to have the right complexity.
Efficient training of small models is particularly important if we do not have large annotated datasets to train big models.
In addition, from practical perspective, to obtain smaller yet accurate models enables deploying fast and accurate models to a limited resource setting \citep{buciluǎ2006model,hinton2015distilling}.

\subsection{Sentiment Analysis}\label{ssec:sa}
Sentiment analysis is an application of NLP to identify the polarity of subjective
information in given source \citep{pang2008opinion}.  
We use the Stanford Sentiment Treebank (SST) \citep{socher2013recursive} an extension of a dataset \citep{pang2005seeing} which has labels for 215,154 phrases in the parse trees of 11,855
sentences sampled from movie reviews.
Real-valued sentiment labels are converted to an integer ordinal label in {0,..,4} by simple thresholding for five classes: very negative, negative, neutral, positive, and very positive.
Therefore the task is posed as a 5-class classification problem.

\begin{table}[t]
\small
\centering
\caption{Classification Accuracies of Training Regimens on Sentiment Analysis Task. The numbers in parantheses are standard deviations. Model gets better at conjunctions if it is trained with Baby-Step curriculum.}
\label{tab:sa-result}
\scalebox{0.85}{
\begin{tabular}{@{}lll@{}}
\toprule
Regimen                   & All &  Conjunctions \\ \midrule
No-CL from \citep{tai2015improved}  & 46.4 (1.1) &   \\
No-CL (our implementation)     & 46.83 (1.1) &    43.88 (1.9)  \\
Curriculum Sorted & 47.42 & 42.88 \\
One-Pass Curriculum      & 45.74 &  43.09      \\
Baby Steps Curriculum     & 47.37  &  46.07    \\ \bottomrule
\end{tabular}
}
\vspace{-10pt}
\end{table}

\noindent \textbf{Dataset Details.} We use the standard
train/dev/test splits of 8544/1101/2210
for the 5-class classification problem.
We flatten the annotated tree structure into sequences of phrases to use finer grained annotations.
We treat the words within the span of an inner phrase as a sequence and use the phrase's annotation as label. 
This results in a bigger training set of 155019 instances.

\noindent \textbf{Experimental Details.}
We follow the previous work \citep{tai2015improved} for the empirical setup.
We use a single layer LSTM with 168 units for the 5-class classification task.
We initialized the word embeddings using 300-dimensional Glove vectors \citep{pennington2014glove} and fine-tuned them during training. 
For optimization, we used RMSprop \citep{dauphin2015rmsprop} with learning rate 0.001 and decay rate of 0.9 with mini-batches of size 128. The patience parameter $p$ for early stopping is 10.

\subsubsection{Results}

As the first step to our more detailed analysis, Table~\ref{tab:sa-result} reports the overall performance of the four learning regimens and the original results stated by \citet{tai2015improved}.
The advantage of CL is most prominent when predicting sentiment for sentences with conjunctions (last column in Table~\ref{tab:sa-result}).
For conjunctions where a span of text contradicts or supports overall sentiment polarity, Baby Step model achieves significantly better results than others. We take a closer look at the LSTM modeling process using a similar probing technique used for the Digit Sum dataset.

\begin{figure}[t]
\centering
\tiny
\centerline{\includegraphics[width=\columnwidth]{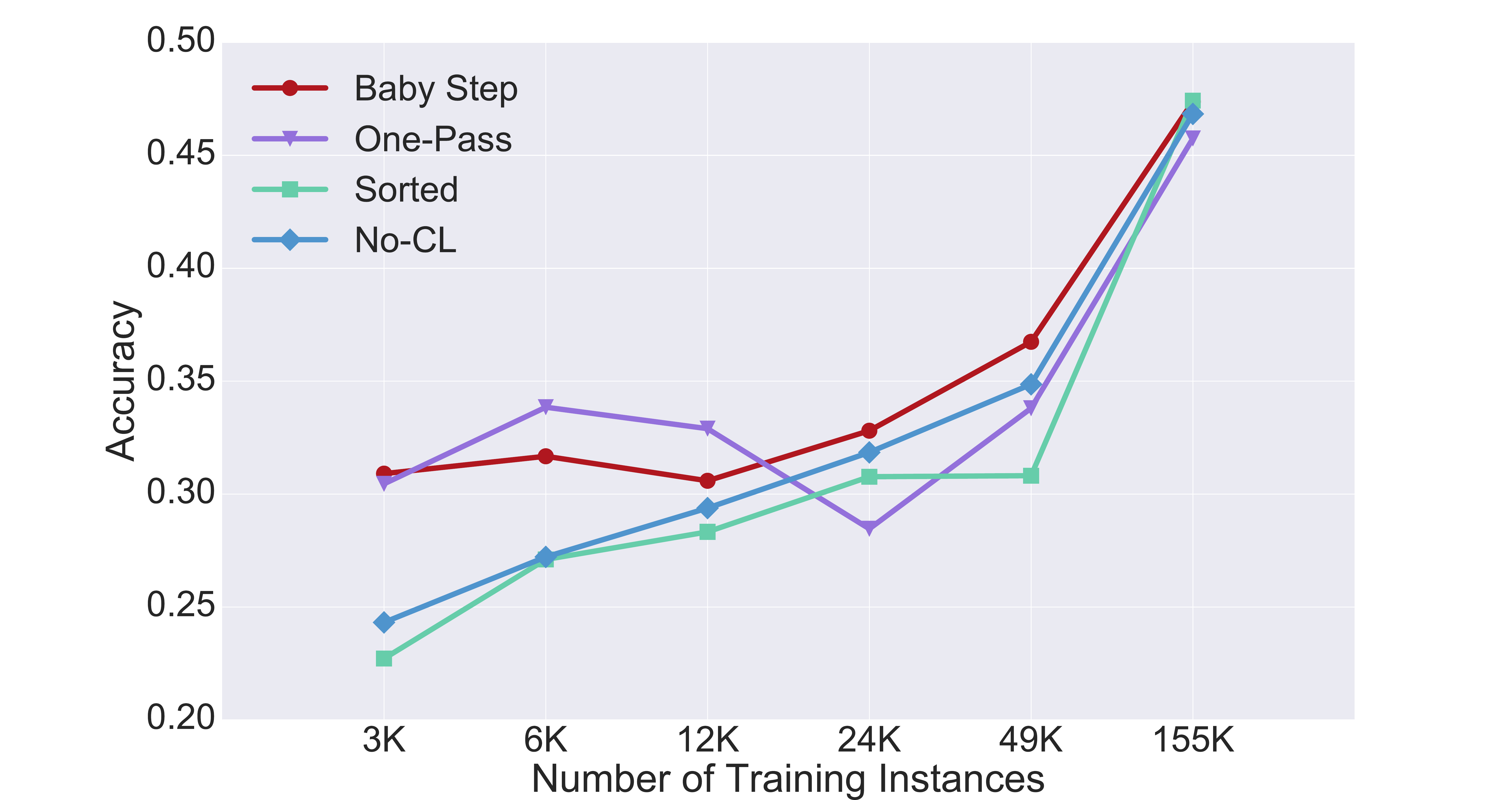}}
\caption{The effect of regimen vs the amount of training data. One-Pass and Baby Steps curriculum regimens gets better results when the training data is limited. They converge to similar points when the amount of data increases.}
\label{fig:sa-data}
\vspace{-10pt}
\end{figure}

\noindent \textbf{Probing Intermediate Representations.} In Figure~\ref{fig:sa-conj}, we qualitatively show how different models process a sentence with a contrastive conjunction originally demonstrated by \citet{socher2013recursive}. 
For each model, we plot the sentiment polarity and the probability of prediction for that polarity after observing a word token.
Unlike the others, Baby Step model changes the sentiment at the appropriate time; after observing ``spice" which constructs a positive statement with the sub-phrase ``but it has just enough spice".
Handling contrastive conjunctions requires a model to merge two conflicting signals (i.e. positive and negative) coming from two directions (i.e. left phrase and right phrase) in an accurate way \cite{socher2013recursive}.
Considering LSTM's limited capacity due to using only signal coming from previous timesteps (i.e processing the sentence from left to right), this result is particularly interesting because Baby Step CL boosts LSTM's performance.

\noindent \textbf{Effect of Training Data Size.} To investigate the role of the amount of training data, we use a varying fraction of training data with learning regimens. Figure~\ref{fig:sa-data} shows the results. CL regimens help when training data is limited.
When the amount of training data increases, the difference between the regimens gets lower.
This result suggests that in low-resource setups, like many of the NLP problems, CL could be useful to improve a model's performance.

\vspace{-5pt}
\section{Conclusion} \label{sec:conclusion}

We examined curriculum learning on two sequence prediction tasks.
Our analyses showed that curriculum learning regimens based on shorter-first approach, help LSTM construct a partial representation of the sequence in a more intuitive way.
We demonstrated that curriculum learning helps smaller models improve performance, 
 contributes more in a low resource setup.
Using a quantitative and qualitative analysis on sentiment analysis, we showed that a model trained with Baby Step curriculum significantly improves for sentences with conjunctions suggesting that curriculum learning helps LSTM learn longer sequences and functional role of the conjunctions.

\newpage
\bibliographystyle{aaai}
\bibliography{00main}
\end{document}